# COEFFICIENTS OF RELATION FOR PROBABILISTIC REASONING

Silvio Ursic
Ursic Computing
810 Ziegler Rd.
Madison, Wisconsin 53714

Definitions and notations with historical references are given for some numerical coefficients commonly used to quantify relations among collections of objects for the purpose of expressing approximate knowledge and probabilistic reasoning.

## 1. INTRODUCTION

Many coefficients that do not correspond to probabilities are being proposed and used for the purpose of considering uncertainty in artificial intelligence. As an example, consider the following assertion: "A car of brand 'A' develops transmission trouble three times more often than with other cars". Using $P(X|Y)$ for the standard Bayesian conditional probability, one might have:

P( car has transmission problems | car brand A ) = .003,
P( car has transmission problems | car not brand A ) = .001,

and we have our "three" as .003 / .001. If this ratio is one, we have statistical independence between the events "car brand A" and "car has transmission trouble". The coefficient "three" in the example above is a factor linking two conditional probabilities. Such factors are widely used in current expert system codes. They are known as "likelihood ratios", "certainty factors", "confidence factors", "evidential factors" (and others), as applied to "weights", "evidence", "beliefs", "rule strengths", "possibilities" (and others), with the purpose of "combine evidence", "maintain truth", "update beliefs", "propagate uncertainties" (and others).

## 2. SOME COEFFICIENTS OF RELATION

A small collection of numerical coefficients has been identified as forming the basic core of coefficients currently being used for uncertain reasoning. They are all old, simple and very useful. What follows does not introduce anything new, or original. Its purpose is *notational* and *computational*. The notation for conditional probabilities is universally accepted (and used). $P(A|B)$ means $P(A\&B) / P(B)$ and not $P(A\&B) / P(A)$ or, say, $P(-A\&B) / P(B)$. Nevertheless, for other coefficients which express some relation between two statistical events distinct from the bayesian ratio, inconsistencies in actual usage remain strong. To be able to use these coefficients fluently, one cannot constantly have to worry if $f(A,B)$ for one implementation is $g(B,A)$, or $h(-A,B)$, or $1 / k(-A,-B)$ for another implementation. Hopefully, what follows should help develop a standard notation and nomenclature for them, which in turn should facilitate their usage.

This note also addresses the problem of making clearer the distinction between what is being computed and the techniques used to compute it.  Perhaps
unintentionally, current efforts with approximate reasoning in artificial
intelligence tend to blend the two subjects together.  The net result of this
state of affairs may have a desirable effect from a marketing point of view.
In the resulting confusion, it is always possible to claim that "our package"
computes something totally different (and much better) than "their package",
with no possibility of being contradicted.  The discussion following the
presentation of [Wise 86] at the 1986 Uncertainty in Artificial Intelligence
Workshop was quite representative of this, and illuminating.  The heated debate
that resulted is a direct consequence of this blending of *problems* with *methods
of solution.*

To reiterate the point, what needs to be computed to simulate probabilistic
aspects of human intelligence is a psychological or perhaps a philosophical
problem.  Coefficients of relation have been found useful for this purpose.
How to carry out the computations specified by these psychological and
philosophical considerations is a problem of a rather distinct nature.  The
specific numerical and combinatorial methods employed to solve some proposed
system of numerical equations will no doubt influence computing times, memory
utilization, problems with numerical stability, precision, etc.  Ultimately,
the available methods of solution determine what can and what cannot be done.
The methods of solution should not, however,  subreptitiously change what is
supposedly being computed.

The definitions and notation for the basic core of coefficients of relation
for reasoning with uncertainty are:

```
|===================================================================|
|   PROBABILITY                                                     |
|   P(A)                        Classical point probability         |
|   P(A|B) = P(A&B) / P(B)      Conditional probability (Bayesian)  |
|                                                                   |
|   INDEPENDENCE (Quetelet)                                         |
|   Q(A|B) = O(A|B) / O(A|-B)   Conditional Odds Ratio              |
|   Q(A:B) = P(A|B) / P(A|-B)   Conditional Probability Ratio       |
|                                                                   |
|   EXCHANGEABILITY (de Finetti)                                    |
|   F(A|B) = O(A|B) / O(B|A)    Conditional Odds Ratio              |
|   F(A:B) = P(A|B) / P(B|A)    Conditional Probability Ratio       |
|===================================================================|
```

We use here "-" for a logical NOT and "&" for a logical AND.  The survey
[Goodman & Kruskal 1954, 1959] will give the reader a taste of the variety of
functions on the four quantities P(A&B), P(A&-B), P(-A&B) and P(-A&-B) that
have been found useful for some purpose.  The paper [Fienberg & Gilbert 1970]
provides interesting geometric insights on independence.  Also see [Good 1965,
1985] for additional stimulating information.

3.  RANGES

Mainly for psychological reasons, it has repeatedly been found necessary to
provide alternate ranges to the  zero-one and zero-infinite ranges of these
coefficients.  In particular, a "minus one to plus one" range has been found
many times very appealing.  The most commonly used ranges and conversions are:

probability range (P-type): zero to one;
odds range       (O-type): zero to infinity;
symmetric range  (S-type): minus one to plus one.

### CONVERSIONS AMONG RANGES

| FROM: | P-type | O-type | S-type |
|---|---|---|---|
| **TO:** | | | |
| P-type | $P$ | $O / (1+O)$ | $(S+1) / 2$ |
| O-type | $P / (1-P)$ | $O$ | $(1+S) / (1-S)$ |
| S-type | $2*P - 1$ | $(O-1) / (O+1)$ | $S$ |

With these ranges, each coefficient can have two additional aliases. For probabilities, however, the symmetric range has been found inconvenient. Similarly, for the independence and exchangeability coefficients, the probability range does not seem useful. The notations for the useful aliases are:

| *MAIN* | | | | *ALTERNATE* | | |
|---|---|---|---|---|---|---|
| P(A) | P-type | [0, 1] | | O(A) | O-type | [0, infinite] |
| P(A\|B) | P-type | [0, 1] | | O(A\|B) | O-type | [0, infinite] |
| Q(A\|B) | O-type | [0, infinite] | | QS(A\|B) | S-type | [-1, +1] |
| Q(A:B) | O-type | [0, infinite] | | QS(A:B) | S-type | [-1, +1] |
| F(A\|B) | O_type | [0, infinite] | | FS(A\|B) | S-type | [-1, +1] |
| F(A:B) | O-type | [0, infinite] | | FS(A:B) | S-type | [-1, +1] |

In general, if $a / b$ is a coefficient of odds type then $a / (a+b)$ is of probability type and $(a-b) / (a+b)$ is of symmetric type. From a computational perspective, alternate ranges are not needed. However, expressing degrees of independence and exchangeability with a coefficient between minus one and one, with a zero indicating independence or exchangeability, seems to sometimes tickle our brain the right way. In many situations we also find that odds are favored over probabilities (see, for example, betting). Perhaps this is how some coefficients of relation are biologically stored, and hence further conversions to the odds scale or to the probability scale involves additional "thinking", which we prefer to avoid. One may classify the situation as one of wanting biologically pleasing units of measure at our disposal.

## 4. PRODUCT PARTITION DEFINITIONS

Let:
P( A & B ) = x,
P( A & -B ) = y,
P( -A & B ) = z,
P( -A & -B ) = w.

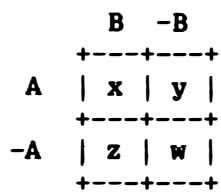

We then have:

*PROBABILITY*
P(A) = 1 - P(-A) = x + y
P(B) = 1 - P(-B) = x + z

*ODDS*
O(A) = 1 / O(-A) = ( x + y ) / ( z + w )
O(B) = 1 / O(-B) = ( x + z ) / ( y + w )

### CONDITIONAL PROBABILITY

$$P(A| B) = 1 - P(-A| B) = x / ( x + z )$$
$$P(A|-B) = 1 - P(-A|-B) = y / ( y + w )$$
$$P(B| A) = 1 - P(-B| A) = x / ( x + y )$$
$$P(B|-A) = 1 - P(-B|-A) = z / ( z + w )$$

### CONDITIONAL ODDS

$$O(A| B) = 1 / O(-A| B) = x / z$$
$$O(A|-B) = 1 / O(-A|-B) = y / w$$
$$O(B| A) = 1 / O(-B| A) = x / y$$
$$O(B|-A) = 1 / O(-B|-A) = z / w$$

### QUETELET PROBABILITIES RATIO

$$Q( A: B) = 1 / Q( A:-B) = ( x / (x+z) ) / ( y / (y+w) )$$
$$Q(-A: B) = 1 / Q(-A:-B) = ( z / (x+z) ) / ( w / (y+w) )$$
$$Q( B: A) = 1 / Q( B:-A) = ( x / (x+y) ) / ( z / (z+w) )$$
$$Q(-B: A) = 1 / Q(-B:-A) = ( y / (x+y) ) / ( w / (z+w) )$$

### QUETELET ODDS RATIO

$$Q(A|B)    = Q(B|A)    = Q(-A|-B)  = Q(-B|-A) =$$
$$1/Q(A|-B) = 1/Q(B|-A) = 1/Q(-A|B) = 1/Q(-B|A) =  xw / yz$$

### de FINETTI PROBABILITIES RATIO

$$F( A: B) = 1 / F( B: A) = ( x + y ) / ( x + z )$$
$$F( A:-B) = 1 / F(-B: A) = ( x + y ) / ( y + w )$$
$$F(-A: B) = 1 / F( B:-A) = ( z + w ) / ( x + z )$$
$$F(-A:-B) = 1 / F(-B:-A) = ( z + w ) / ( y + w )$$

### de FINETTI ODDS RATIO

$$F( A|B) = F(-B|-A) = 1 / F(B| A) = 1 / F(-A|-B) = y / z$$
$$F(-A|B) = F( B|-A) = 1 / F(B|-A) = 1 / F(-A| B) = x / w$$

Notice that we have four distinct coefficients of type Q(A:B),  F(A:B) and P(A|B), two of type F(A|B) and only one of type Q(A|B). From a computational perspective, Q(A|B) and F(A|B) are both cheaper than Q(A:B) and F(A:B). They are the preferred form for independence and exchangeability. The independence of A and B can be declared both with Q(A|B) = 1 and   Q(A:B) = 1. Similarly, the exchangeability of A and B can be set with F(A|B) = 1 or F(A:B) = 1. For values near one, the pair Q(A|B) and Q(A:B) and also the pair F(A|B) and F(A:B) convey almost the same information.

## 5. CONDITIONAL PROBABILITY

Thanks to [Bayes 1763], the notation P(A|B) is a universal standard for conditional probability.

## 6. INDEPENDENCE

The papers [YULE 1900, 1903, 1912] introduced the "Q" notation for these two independence coefficients. Yule choose "Q" in honor of Quetelet. The book translation from the French [Quetelet 1849] is one of the few references given by Boole as having had an impact on his writings in probability. The deciphering of the very verbose exposition in Quetelet book might have been Boole's motivation to develop the compact notation that we today know as "Boolean algebra". with the heavy emphasis on probabilities he gave to his work. In connection with independence of statistical events, the paper [Wilbraham 1854] can be considered a classic in misinterpretation of its applicability. In general, one does not know if two given events are independent. By default their relation is unknown. This limited role assumed by statistical independence has been a stumbling block to the applicability of existing statistical results for artificial intelligence and uncertain reasoning. A second reason for the unwarranted distaste shown by some researchers in AI towards long established ideas and results in probability can be traced to plain and simple "algorithmic draught". The computational problems that arise when we wish to implement ideas and principles exemplified with three or four events to three or four thousand events necessitate the development of entirely new algorithmic approaches.

## 7. EXCHANGEABILITY

Exchangeability introduces many properties that mirror the properties of independence. For example, with N exchangeable events we have N degrees of freedom among the $2^N$ product partition events, exactly as with independent events. Consult [de Finetti 1937, 1969], [Chrisma 1971, 1982], [Diaconis 82], [Galambos 82] to get acquainted with results and a growing bibliography on exchangeability. Many statistical events in approximate reasoning are more precisely modeled by assuming exchangeability than independence. The constraints introduced by exchangeability blend easily with the polyhedral constraints arising from the fact that the statistical events under consideration are at the onset defined with arbitrary boolean functions, which are also polyhedral constraints. Consult [Ursic 1987] for additional information on the constraints on the probabilities of the product partition arising from boolean constraints.

## 8. CONCLUDING REMARKS

It seems inevitable that many other coefficients, besides the ones reported here, will become necessary to quantify finer and finer properties for relations among events. For example, chi-square for two events is also an independence coefficient. However, the coefficients $Q(A|B)$, $Q(A:B)$, $F(A|B)$ and $F(A:B)$ stand out for their simplicity and generality.

In using these coefficients for approximate reasoning one tacitly assumes the validity of probabilistic ideas as a framework for reasoning with uncertainty. Insights on this can be found in [Cheeseman 1985]. A further consequence of the probabilistic approach is the demise of the concept of a logical deduction. All what we truly have is a system of linear and nonlinear equations, some derived from coefficients of relation and some derived from boolean definitions of events. As an example of this, consult [Dempster 1967], [Quinlan 1983, 1985]. The sought answers, in the form of probabilities of some events and of coefficients of relation for some collections of events, are simply obtained by solving appropriate systems of algebraic equations. If possible, we solve them by analytical means. If analytical methods fail, as has been the case with many probabilistic problems arising from artificial intelligence, we solve them with numerical methods. If exact solutions cannot be obtained in reasonable amounts of time, we use approximate methods.

Describing Gaussian elimination or the simplex method as procedures performed by an inference net, or by an algorithm performing a deduction, or by a forward or backward chain of inferences does not seem to describe the methods of solution that can be actually employed. Efficient methods of solution for these systems of equations may have very little relation with what they mean to the end user. As a point fo reference, consider the fast Fourier transform algorithm and its relation to some intuitive meaning of harmonic analysis. The point of view taken here and in [Ursic 87] is that most (if not all) of what is currently being proposed as reasoning with uncertainty can be interpreted as consisting of the problem of solving systems of linear and nonlinear inequalities and equations with the unknowns being some sought collection of coefficients of relation. Advances will therefore come from two sources. First, the equations to be solved must be clearly and precisely stated. This is not an easy task, especially with respect to the constraints arising from logical conditions. Second, we systematically develop specialized numerical methods for the solution of the particular systems of equations so obtained. As a consequence, with the help of standardized test problems, available methods of solution, trade-offs between precision and computing times, specialized sub-problems, etc., can be analyzed and compared.

From this perspective, most (if not all) of the current efforts in approximate reasoning can be interpreted as being ad hock methods of solution (neither very efficient nor very precise) for the very special systems of equations and inequalities arising from the logical and statistical constraints defining the problem at hand. Trying to mimic what we perceive as being the solution methods employed by biological computers for the problem may be counterproductive. Biological systems have such severe limitations in energy consumption and energy density that the solutions they developed to the problems we are facing may not be suitable to the tools at our disposal. One should consider that we do not build airplanes with flapping wings.